\documentclass[lettersize,journal]{IEEEtran}
\usepackage{amsmath,amsfonts}

\usepackage{algorithm}
\usepackage{algpseudocode}
\usepackage{array}
\usepackage[caption=false,font=normalsize,labelfont=sf,textfont=sf]{subfig}
\usepackage{textcomp}
\usepackage{stfloats}
\usepackage{url}
\usepackage{verbatim}
\usepackage{graphicx}
\usepackage{multirow}
\usepackage{flushend}
\usepackage{subcaption}
\usepackage{cite}
\usepackage{tabularx}
\usepackage{xcolor}
\usepackage{etoolbox}
\usepackage{caption}
\usepackage{colortbl}
\usepackage[hidelinks]{hyperref}
\IEEEoverridecommandlockouts

\newcommand{\insertfig}
{\includegraphics[width=\linewidth, trim=0.0in 0.0in 0.0in -1.0in, clip]{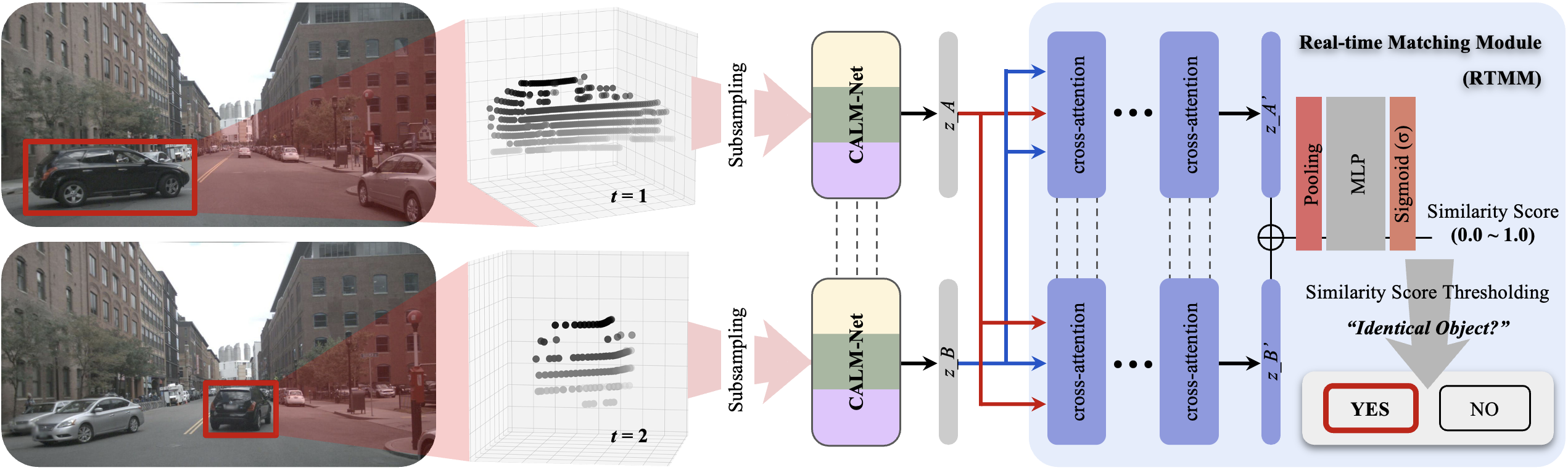}
\captionof{figure}{\textbf{The re-identification process using real-time matching module (RTMM).} Input point clouds are subsampled and passed toward the siamese neural network, exemplified here by CALM-Net. The extracted features are then processed through a sequence of cross-attention blocks within RTMM, producing a similarity score. Finally, the similarity score is thresholded to predict whether two input measurements correspond to the identical object or not.}
\vspace{-15px}
}

\makeatletter
\apptocmd{\@maketitle}{\centering\insertfig}{}{}
\makeatother

\title{CALM-Net: Curvature-Aware LiDAR Point Cloud-based Multi-Branch Neural Network for Vehicle Re-Identification}
\author{Dongwook Lee, Sol Han, Jinwhan Kim
    \thanks{The authors are with the Robotics Program (e-mail: dongwooklee1201@kaist.ac.kr) and Department of Mechanical Engineering (e-mail: dream4future@kaist.ac.kr; jinwhan@kaist.ac.kr), Korea Advanced Institute of Science and Technology, Daejeon 34141, South Korea. \textit{(Corresponding author : Jinwhan Kim)}}
}

\begin{document}

\maketitle
\thispagestyle{empty}
\pagestyle{empty}





\begin{abstract}
This paper presents CALM-Net, a curvature-aware LiDAR point cloud-based multi-branch neural network for vehicle re-identification. The proposed model addresses the challenge of learning discriminative and complementary features from three-dimensional point clouds to distinguish between vehicles. CALM-Net employs a multi-branch architecture that integrates edge convolution, point attention, and a curvature embedding that characterizes local surface variation in point clouds. By combining these mechanisms, the model learns richer geometric and contextual features that are well suited for the re-identification task. Experimental evaluation on the large-scale nuScenes dataset demonstrates that CALM-Net achieves a mean re-identification accuracy improvement of approximately 1.97\% points compared with the strongest baseline in our study. The results confirms the effectiveness of incorporating curvature information into deep learning architectures and highlight the benefit of multi-branch feature learning for LiDAR point cloud-based vehicle re-identification. The source code of the CALM-Net and the Re-ID evaluation can be found here: \href{https://github.com/ldw200012/CALM-Net.git}{\textcolor{blue}{https://github.com/ldw200012/CALM-Net.git}}
\end{abstract}


\section{Introduction}
\IEEEPARstart{V}{ehicle} re-identification (Re-ID) plays a pivotal role in intelligent transport systems, supporting cross-camera tracking, traffic analysis, and the safety of autonomous driving. By linking observations of the same vehicle across time and space, Re-ID enhances multi-object tracking by addressing failures of motion-only models under occlusion, trajectory fragmentation, or identity switches \cite{3dmot}. In such cases, motion-based tracking alone may mistakenly transfer a trajectory to the wrong vehicle, whereas Re-ID provides complementary appearance and geometry-based cues that reconnect the correct track. Beyond its contribution to robust tracking, Re-ID is also crucial for broader applications such as traffic surveillance, security monitoring, congestion management, and mobility analytics \cite{varid, knowledge-mb, semantic-reid}.

While cameras have traditionally dominated vehicle Re-ID research due to their rich RGB information \cite{video-surveil, large-surveil, veri-wild}, they remain sensitive to illumination changes, occlusions, and varying viewpoints. Consequently, a large body of work has been devoted to camera-based vehicle Re-ID, where researchers design sophisticated architectures to overcome core computer vision challenges such as pose variation, background clutter, and the high similarity among vehicle appearances. These efforts include viewpoint-aware learning \cite{varid}, multi-branch feature fusion \cite{knowledge-mb, mb-enhanced, tbe-net}, transformer-based attention models \cite{semantic-reid, mskat, mart, multimodal-transformer}, and domain adaptation strategies \cite{multimodal-transformer, mask-aware, strdan, i2ida}. Despite such progress, performance still degrades significantly under drastic viewpoint shifts and poor visual conditions.

In contrast, point cloud data provides accurate 3D geometric information, scale invariance, and robustness to lighting, making it a promising modality for reliable Re-ID in complex urban environments \cite{freeperson, radar-reid, lidar-reid}. Recent studies have begun to explore point cloud-based Re-ID, showing that high-resolution point clouds can even surpass image-based methods in certain scenarios. However, despite this promise, LiDAR point cloud-based vehicle Re-ID remains underexplored, particularly with respect to leveraging intrinsic geometric descriptors that capture fine-grained surface variation and are inherently viewpoint-robust and discriminative.

To address this gap, we propose curvature-aware LiDAR point cloud-based multi-branch neural network (CALM-Net). CALM-Net integrates three complementary feature extraction mechanisms: (1) edge convolution to model local topology, (2) point attention to capture contextual dependencies, and (3) curvature embedding to encode geometric surface variation. By combining these branches, our model learns discriminative, geometry-driven vehicle embeddings that are resilient to viewpoint and environmental changes. The main contributions of this paper are as follows:

\begin{itemize}
    \item We propose \textbf{CALM-Net}, the first LiDAR point cloud-based vehicle Re-ID network that explicitly incorporates curvature information through a learnable \textit{curvature embedding}, enabling fine-grained geometric reasoning under viewpoint changes and sparse measurements.
    
    \item We design a \textbf{multi-branch neural architecture} that integrates edge convolution, point attention, and curvature embedding, allowing the model to capture complementary local, contextual, and structural features from 3D point clouds.

    \item We introduce a \textbf{hybrid point subsampling strategy}—random sampling during training and farthest point sampling (FPS) during inference—that improves generalization and structural consistency, and we validate its effectiveness across multiple object classes included in a large-scale nuScenes dataset.
\end{itemize}

\section{Related Works}
\subsection{Vehicle Re-ID}

Vehicle Re-ID has been extensively studied in the computer vision community, primarily with camera-based data, where numerous methods have been proposed to overcome challenges such as viewpoint variation, illumination changes, and occlusion. To reduce the reliance on large-scale labeled datasets, a number of studies investigate unsupervised and domain adaptation approaches. For example, multimodality adaptive transformers with mutual learning have been proposed for unsupervised adaptation \cite{multimodal-transformer}, inter- and intra-cluster reorganization has been used to refine pseudo-labels \cite{inter-intra}, mask-aware pseudo label denoising was introduced to suppress noise during self-training \cite{mask-aware}, and bi-level semantic augmentation in feature space has been presented to improve robustness \cite{bilevel}.

Other efforts employed adversarial learning and style-irrelevant disentanglement to enhance cross-domain performance. Since vehicles exhibit large appearance changes across different viewpoints, several works have focused on viewpoint-aware modeling. A viewpoint-aware triplet loss (VARID) was introduced in \cite{varid}, UAV-based Re-ID was addressed through posture calibration and cross-view metric learning in \cite{uav}, an identity-unrelated decoupling model was proposed to separate viewpoint and background factors in \cite{id-unrelated}, and a mask-aware reasoning transformer was developed to improve robustness under occlusion and perspective changes \cite{mart}.

In parallel, multi-branch networks have demonstrated their effectiveness in extracting complementary local and global cues, including a knowledge-driven multi-branch interaction network \cite{knowledge-mb}, TBE-Net with three branches and part-aware pooling \cite{tbe-net}, and a multi-branch enhanced discriminative network \cite{mb-enhanced}, while other approaches such as quadruple-directional pooling also highlight the benefits of decomposing vehicle features into finer components \cite{quadruple}. More recently, transformers have gained traction in vehicle Re-ID by capturing long-range dependencies and semantic cues, with methods such as semantic-oriented feature coupling transformer \cite{semantic-reid}, multi-scale knowledge-aware transformer \cite{mskat}, and mask-aware reasoning transformer \cite{mart}. Graph-based methods have also been explored, including hybrid pyramidal graph networks to capture spatial significance \cite{pyramidal} and progressive fusion graph frameworks for multi-modal Re-ID \cite{grah-multimodal}.

Beyond visual-only approaches, cross-modal and attribute-guided frameworks have been studied to incorporate richer priors, such as unbiased causal feature enhancement \cite{unbiased}, disentanglement of identity-unrelated factors at both image and feature levels \cite{disentangle}, text-to-image vehicle Re-ID with a unified benchmark \cite{t2i}, and deep hashing for efficient large-scale retrieval \cite{hashing}. Finally, practical systems have been proposed for deployment in real traffic networks, including traffic-informed multi-camera sensing systems integrating Re-ID with spatio-temporal graph inference \cite{tims} and dual domain multi-task frameworks for joint identity and attribute recognition \cite{dualdomain}. Despite this broad progress, the majority of existing methods remain grounded in camera-based representations, leaving LiDAR point cloud-based vehicle Re-ID relatively unexplored.

\subsection{Point Cloud-based Re-ID}

Research on point cloud-based Re-ID has progressed along two main directions: \emph{person Re-ID} and \emph{vehicle Re-ID}.  

Early efforts primarily focused on person Re-ID, often using synthetic point clouds derived from 2D images with added depth information. These methods projected 2D images into estimated 3D point clouds and evaluated performance on large-scale person Re-ID datasets such as Market-1501, DukeMTMC-reID, MSMT-17, and CUHK03-NP \cite{32,33,34,35,36,37}. Other works generated pseudo point clouds by combining Kinect depth with RGB data \cite{38}, while approaches using RGB-D cameras with skeletal tracking enabled real-time 3D reconstruction of freely moving persons for Re-ID \cite{freeperson}. More recently, ReID3D introduced the first LiDAR point cloud-based person Re-ID model, leveraging a graph-based encoder to extract 3D body features, trained first on a simulated dataset of 600 individuals and later on real-world LiDAR scans of 320 people in outdoor environments \cite{39}.

\begin{figure*}[!ht]
    \centerline{\includegraphics[width=1.0\linewidth]{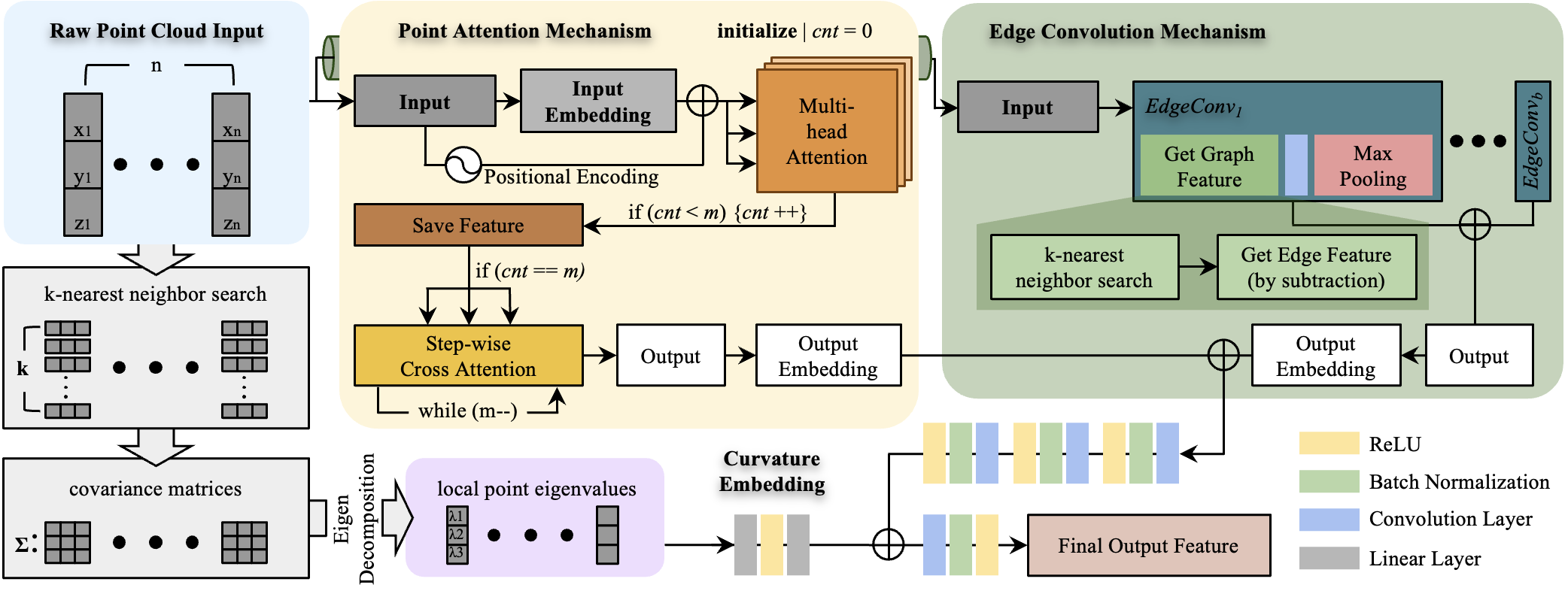}}
    \caption[The structure of the proposed CALM-Net network]
    {\textbf{The structure of the proposed CALM-Net network.} The input LiDAR point cloud is processed through edge convolution and point attention modules, while curvature embedding is encoded from local point eigenvalues. The output features are aggregated into a unified embedding for vehicle Re-ID.}
    \label{fig_calm_horizontal}
\end{figure*}

Beyond persons, researchers have also begun to investigate vehicle Re-ID using point clouds. Radar-based person Re-ID demonstrated high accuracy in small cohorts, highlighting radar’s robustness and cost-effectiveness \cite{radar-reid}, but the superior resolution and range of LiDAR make it particularly suitable for vehicle-scale re-ID in complex traffic scenes. One line of work connected LiDAR data to multi-object tracking frameworks, where point clouds were projected into bird’s-eye-view (BEV) images for vehicle association \cite{40}. However, BEV projection discards rich 3D morphological cues. More recently, a large-scale study explicitly addressed multi-class object re-ID—including both vehicles and pedestrians—using annotated LiDAR data from nuScenes and Waymo \cite{lidar-reid, 41,42}. By introducing a real-time matching module (RTMM) within a siamese architecture, the study showed that higher point cloud density enables LiDAR point cloud-based vehicle re-ID to surpass camera-based methods, underscoring the robustness of 3D geometric features in challenging urban environments \cite{lidar-reid}.

\section{Method}

CALM-Net integrates three complementary branches—edge convolution, point attention, and curvature embedding—into a unified multi-branch framework. The rationale behind this design is that each branch captures different aspects of the data: edge convolution encodes local topology, point attention emphasizes global context, and curvature features provide viewpoint-invariant surface variation. The overall model structure of CALM-Net is illustrated in Fig. \ref{fig_calm_horizontal}.

\subsection{Curvature Embedding} \label{subsec_curvature}

In 3D space, the eigenvalues of a covariance matrix have a geometric interpretation related to scaling and direction. For the covariance matrix of a set of 3D points, these eigenvalues describe the point distribution, indicating both the orientation and magnitude of the spread. In mobile robotics, this concept is frequently applied to determine the heading direction of a vehicle. For example, cars, with their elongated shape, tend to have their largest variance along the longitudinal axis. Thus, the eigenvector corresponding to the largest eigenvalue, representing maximum variance, can be used to estimate the vehicle’s heading.

We extend this geometric concept to smaller, localized regions in order to capture finer point surface variation. Instead of calculating the eigenvalues for an entire object’s point cloud, we perform a $k$-nearest neighbor search centered on a specific point. This localized approach ensures that the eigenvalues characterize the point distribution within a confined neighborhood. By selecting an appropriate value for $k$, one can infer whether the local patch is more planar, edge-like, or highly curved—a property we define as \textit{curvature information}. Figure \ref{fig_curvature_ext} visually illustrates how eigenvalues reflect local curvature, and Eqns. \ref{eq:eigenvalue}--\ref{eq:eigendecomposition} detail the mathematical process for computing the 3D eigenvalues from the point cloud. \(X(i)\) denotes the set of $k$-nearest points around the point $x_i$ (including $x_i$ itself) in the following equations:

\begin{figure}[h]
    \centerline{\includegraphics[width=1.0\linewidth]{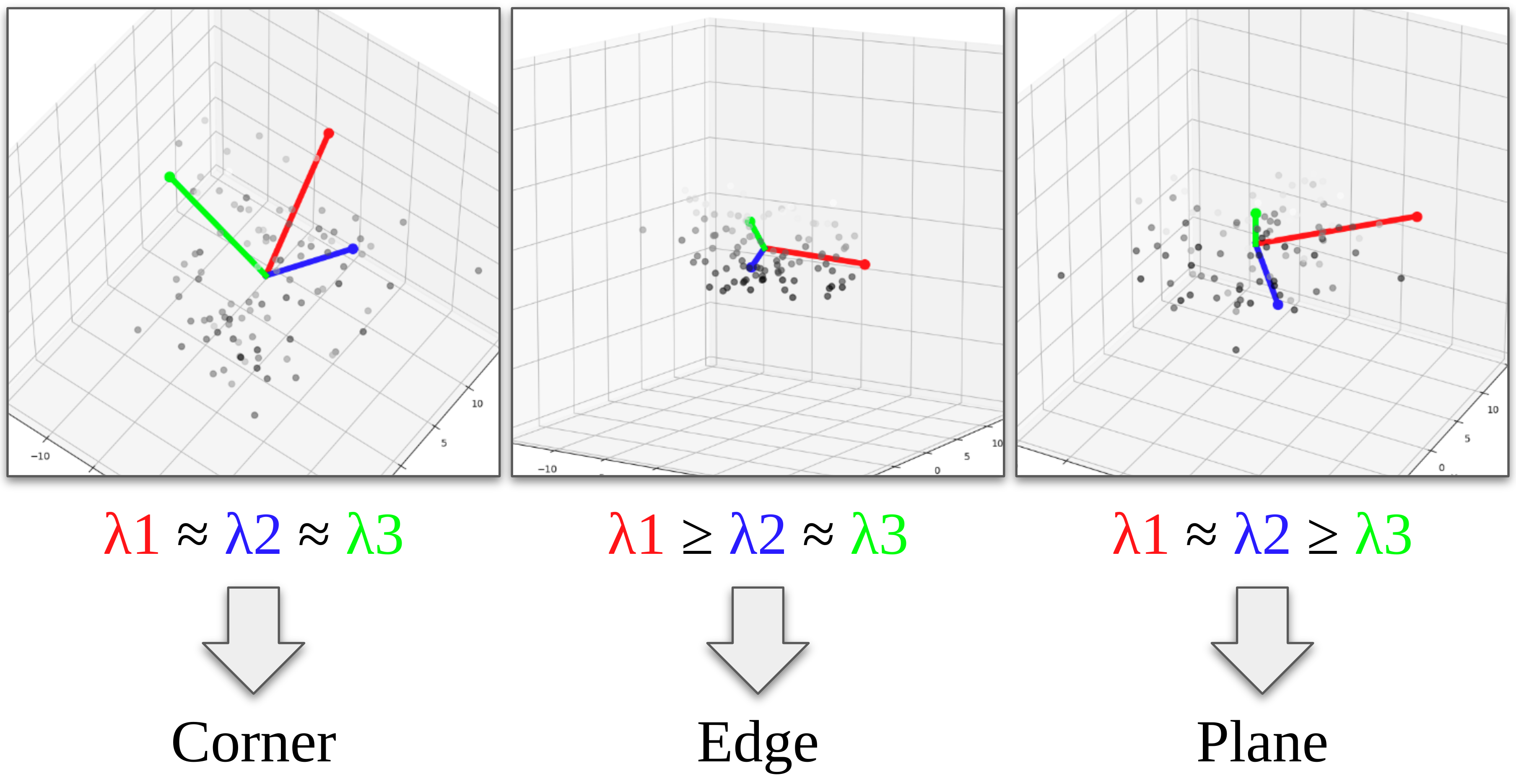}}
    \caption[Visual explanation of curvature information]
    {\textbf{Visual explanation of curvature information.} The relationship between the 3D eigenvalues of a point's local covariance matrix reveals the curvature information of the point surface. The red, blue, and green vectors represent eigenvectors, with their lengths proportional to the corresponding eigenvalues.}
    \label{fig_curvature_ext}
\end{figure}

\begin{figure}[h]
    \centerline{\includegraphics[width=1.0\linewidth]{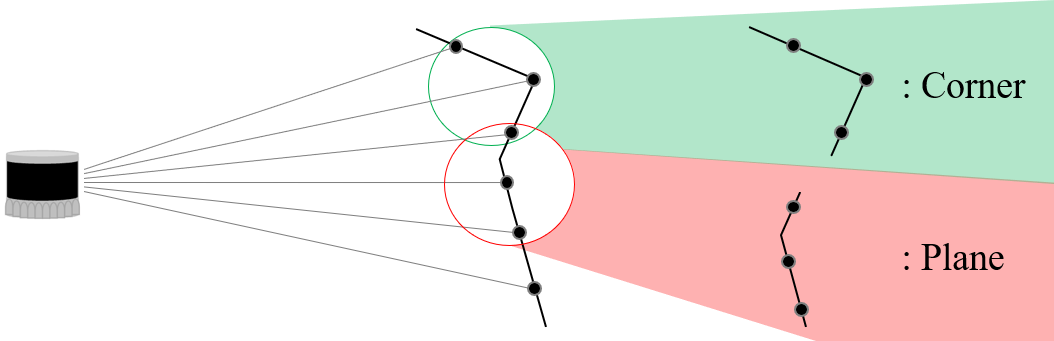}}
    \caption[Inaccurate curvature interpretation due to LiDAR sparsity]
    {\textbf{Inaccurate curvature interpretation due to LiDAR sparsity.} Due to the fixed angular resolution of LiDAR, certain regions of a 3D object may be misinterpreted with incorrect curvature information.}
    \label{fig_lidar_curvature}
\end{figure}

\vspace{-15px}
\begin{equation}\label{eq:eigenvalue}
x_i^c = \sum_{x_j \in X(i)} \frac{x_j}{k}, \quad k = \text{number of k-nearest points},
\end{equation}

\begin{equation}
M_i = \frac{1}{k} \sum_{x_j \in X(i)} (x_j - x_i^c)(x_j - x_i^c)^T,
\end{equation}

\begin{equation}\label{eq:eigendecomposition}
M_i = V_i \Lambda_i V_i^T,
\end{equation}
\vspace{-5px}

\noindent
where the diagonal values of $\Lambda_i$ represent the eigenvalue set $\lambda_i^1, \lambda_i^2, \lambda_i^3$ for point $x_i$ in descending order, providing insight into the degree of local surface variation. However, directly interpreting curvature by thresholding these eigenvalues is challenging, particularly in large datasets with diverse objects. Moreover, the inherent sparsity of LiDAR data exacerbates this issue, as demonstrated in Fig. \ref{fig_lidar_curvature}, where the limited resolution of LiDAR measurements can lead to inaccurate curvature interpretation.

To address this challenge, we encode the eigenvalues into a higher-dimensional feature vector, capturing both linear and non-linear relationships among them. We refer to this as the \textit{curvature embedding}. This descriptor not only reflects the linear relationships associated with the eigenvalues of a local point patch but also reveals significant non-linear interactions, providing a more comprehensive representation of surface variation compared to raw $xyz$ coordinates. We utilize two linear layers with a non-linear activation function (ReLU) between them to encode the eigenvalues, as shown in Eqn. \ref{eq:eigen_layers}, where $\phi_1$ and $\phi_2$ represent linear transformations and ReLU introduces non-linearity, ultimately producing the final curvature embedding:

\begin{equation}\label{eq:eigen_layers}
\text{CurvEmbed}(\Lambda) = \phi_2(\text{ReLU}(\phi_1([\lambda_1, \lambda_2, \lambda_3]))).
\end{equation}

\subsection{Edge Convolution and Point Attention Multi-Branch}

The raw LiDAR point cloud branch of CALM-Net is designed to capture complementary information through two mechanisms: edge convolution, which focuses on local neighborhood geometry, and point attention, which emphasizes long-range contextual dependencies. These two mechanisms are processed in parallel and their outputs are aggregated, enabling the model to leverage both fine-grained structural cues and global salience.  

\subsubsection{Edge Convolution (EC)}  
Edge convolution captures local geometric relationships by modeling the transformation between each point and its neighbors. Given a point cloud $X = \{x_i \in \mathbb{R}^3\}_{i=1}^n$, for each point $x_i$ we define its $k$-nearest neighbors as $\mathcal{N}(i)$. The edge feature between $x_i$ and $x_j \in \mathcal{N}(i)$ is computed as:  

\begin{equation}\label{eq:edgeconv}
h_\theta(x_i, x_j) = \text{ReLU}\left( \theta \cdot (x_j - x_i) + \phi \cdot x_i \right),
\end{equation}
\noindent
where $\theta$ and $\phi$ are learnable weight matrices. The aggregated edge convolution feature for $x_i$ is then obtained as: 

\begin{equation}\label{eq:EC}
\text{EC}(x_i) = \max_{x_j \in \mathcal{N}(i)} h_\theta(x_i, x_j).
\end{equation}

This formulation ensures that local geometric variation, such as surface orientation and neighborhood topology, is encoded into the point representation.  

\vspace{10px}
\subsubsection{Point Attention (PA)}  
While edge convolution focuses on local geometry, point attention highlights contextual dependencies across the entire point cloud. Inspired by transformer-based attention, each point feature is projected into query ($Q$), key ($K$), and value ($V$) spaces:  

\begin{equation}
Q = XW_Q, \quad K = XW_K, \quad V = XW_V,
\end{equation}
\noindent
where $W_Q$, $W_K$, and $W_V$ are learnable matrices. The attention weight between point $i$ and point $j$ is then computed as:  
\begin{equation}
\alpha_{ij} = \frac{\exp\left( Q_i K_j^T / \sqrt{d} \right)}{\sum_{l=1}^n \exp\left( Q_i K_l^T / \sqrt{d} \right)},
\end{equation}
\noindent
where $d$ is the feature dimension. The updated representation of point $i$ is obtained as a weighted sum:  

\begin{equation}\label{eq:PA}
\text{PA}(x_i) = \sum_{j=1}^n \alpha_{ij} V_j.
\end{equation}

This allows each point to selectively attend to other points, capturing long-range contextual salience and discriminative cues beyond its local neighborhood.

\subsection{Overview of CALM-Net Structure}
As illustrated in Fig. \ref{fig_calm_horizontal}, the raw point cloud is processed through the edge convolution (EC) and point attention (PA) modules in parallel, producing two complementary embeddings that are aggregated into a single representation $\text{B1}(X)$. In parallel, curvature information is derived and passed through two linear layers and ReLU, yielding the curvature embedding $\text{CurvEmbed}(\Lambda)$.

Then, $\text{B1}(X)$ and $\text{CurvEmbed}(\Lambda)$ are concatenated and passed through convolution (Conv) and batch normalization (BN) to form $\text{B2}(X)$. Final embedding $\text{CALM-Net}(X)$ is produced by passing $\text{B2}(X)$ into ReLU layer. The process can be expressed as:

\begin{equation}\label{eq:CALM_net}
    \text{CALM-Net}(X) = \text{ReLU}\big(\text{B2}(X) \big),
\end{equation}

\begin{equation}\label{eq:B2}
    \text{B2}(X) = \text{BN}\Big(\text{Conv}\big(\text{B1}(X) \oplus \text{CurvEmbed}(\Lambda) \big)\Big),
\end{equation}
\vspace{-10px}

\begin{equation}\label{eq:B1}
    \text{B1}(X) = \text{MLP}_{\text{conv}}\Big( \text{PA}(X) \oplus \text{EC}(X) \Big),
\end{equation}
\vspace{-1px}

\noindent
where $\oplus$ denotes concatenation. This multi-branch aggregation enables CALM-Net to integrate heterogeneous geometric and contextual cues.

\subsection{Input Point Subsampling Strategy}

The subsampling of input points plays a critical role in the performance of point cloud-based Re-ID models. The way points are subsampled directly affects how well a network learns discriminative features and generalizes to unseen examples. Previous works, such as RTMM, have employed random subsampling as a straightforward and computationally efficient approach.  

An alternative is FPS, which iteratively selects points such that each new point is the farthest from those already sampled. By construction, FPS preserves the global distribution of the input cloud, ensuring more uniform coverage and fidelity to the object’s morphology. This makes FPS particularly advantageous for inference tasks where geometric consistency is critical.  

Each method has unique advantages and limitations. Random subsampling introduces stochasticity into the training process, acting as a form of data augmentation analogous to random cropping in image analysis. This randomness encourages robustness by preventing the model from overfitting to specific local arrangements. In contrast, FPS emphasizes structural integrity by capturing uniform surface coverage, which enhances shape preservation but does not contribute variability during training.  

\begin{figure}[!h]
    \centerline{\includegraphics[width=1.0\linewidth]{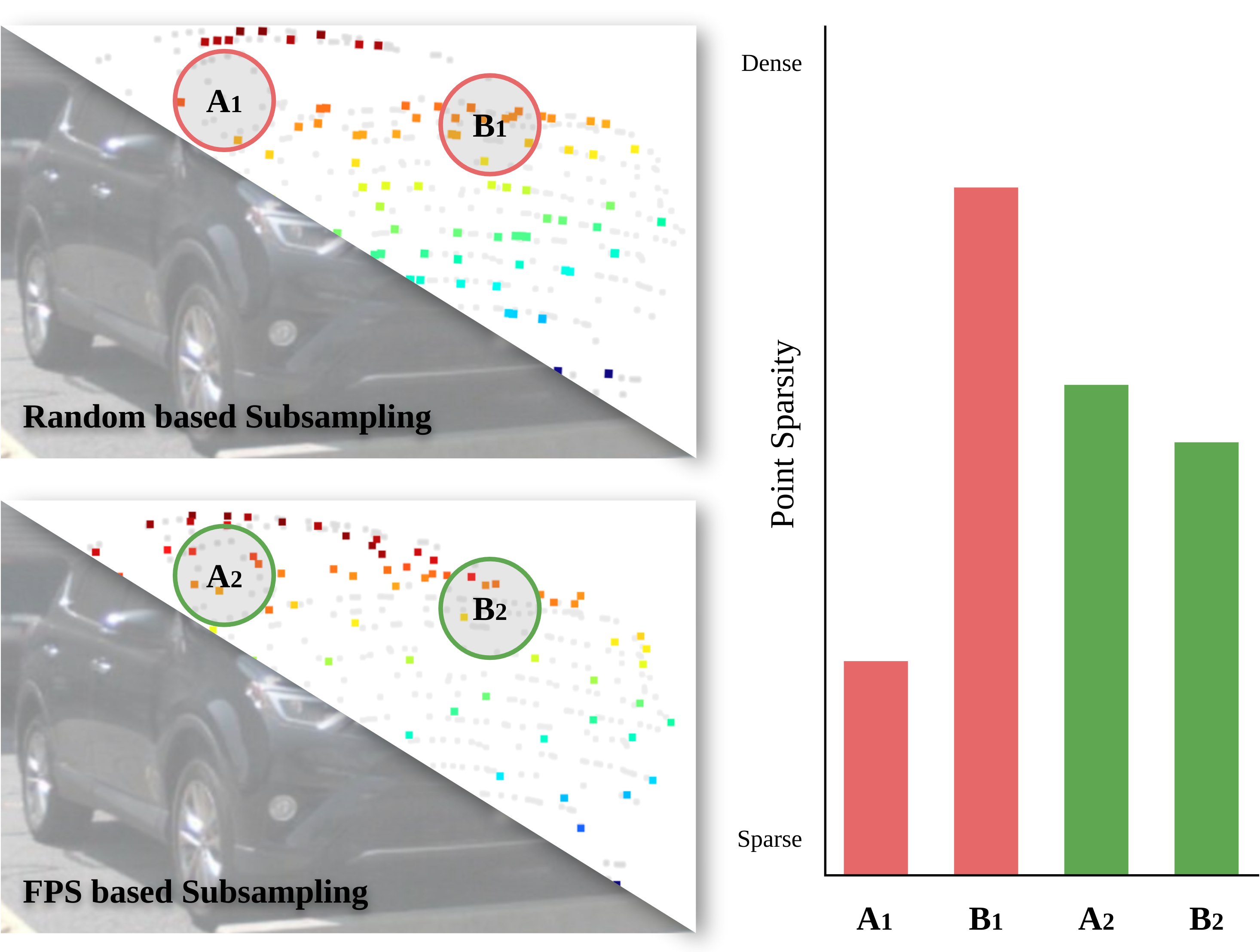}}
    \caption[Comparison of random and FPS-based subsampling]
    {\textbf{Comparison of random and FPS-based subsampling.} Random subsampling introduces irregular densities within local patches, whereas FPS produces more uniform coverage across the object surface.}
    \label{fig_randomfps}
\end{figure}

To balance these complementary properties, we adopt a hybrid subsampling strategy: random subsampling during the training stage to improve robustness through stochastic augmentation, and FPS during inference to ensure uniform coverage and high-fidelity preservation of the vehicle’s geometry. Figure \ref{fig_randomfps} compares the two approaches, illustrating how random subsampling introduces irregular density distributions while FPS achieves more even point spacing across the object surface.  

\subsection{Training Objective}

The training objective is formulated as a binary cross-entropy loss, which determines whether two input point clouds correspond to the same identity. Given predictions \(x_i\) (with 1 indicating the same identity and 0 otherwise) and ground truth labels \(y_i\), the loss is defined as:

\begin{equation}\label{eq:binary_entropy}
L(x, y) = \frac{1}{n} \sum\limits_{i=1}^{n} \left( y_i \log(x_i) + (1 - y_i) \log(1 - x_i) \right),
\end{equation}
\noindent
where \(n\) denotes the number of samples. This objective encourages the model to maximize confidence for correct matches while penalizing incorrect predictions, thereby driving the learning of discriminative embeddings for vehicle Re-ID.  

\section{Experiment} \label{sec_experiment}

\subsection{Experiment Settings}

To evaluate vehicle Re-ID performance, we report three metrics: accuracy, F1 positive score, and F1 negative score. During evaluation, we construct pairs of measurements from the validation set. If both measurements originate from the same object, the pair is labeled as a \textit{positive pair}; otherwise, it is labeled as a \textit{negative pair}. The model predicts whether each pair corresponds to the same identity. Comparing predictions with ground truth yields counts of true positives (TP), true negatives (TN), false positives (FP), and false negatives (FN), from which accuracy and the two class-specific F1 scores are computed. This pairwise protocol directly reflects the operational setting of vehicle Re-ID, where the goal is to decide whether two observations belong to the same vehicle. 

The architectural parameter details of CALM-Net are summarized in Table~\ref{tab_calm_params}.

\vspace{4px}
\begin{table}[h]
\caption[CALM-Net Model Parameter Details]
{\textbf{CALM-Net Model Parameter Details}}
\centering
\label{tab_calm_params}
\setlength{\tabcolsep}{20pt}  
\renewcommand{\arraystretch}{1.5}  
\begin{tabularx}{\columnwidth}{l|l}
\hline 
Parameter & Value\\
\hline \hline
Input Shape & $n \times 3$ \\
\hline
PA($\cdot$) Output Feature Shape & $n \times 128$ \\
\hline
EC($\cdot$) Output Feature Shape & $n \times 32$ \\
\hline
CurvEmbed($\cdot$) Output Feature Shape & $n \times 16$ \\
\hline
CALM-Net($\cdot$) Output Feature Shape & $n \times 128$ \\
\hline
$k$ value for k-NN & 10 \\
\hline
\end{tabularx}
\end{table}
\vspace{-10px}

\begin{figure*}[!ht]
    \centerline{\includegraphics[width=1.0\linewidth]{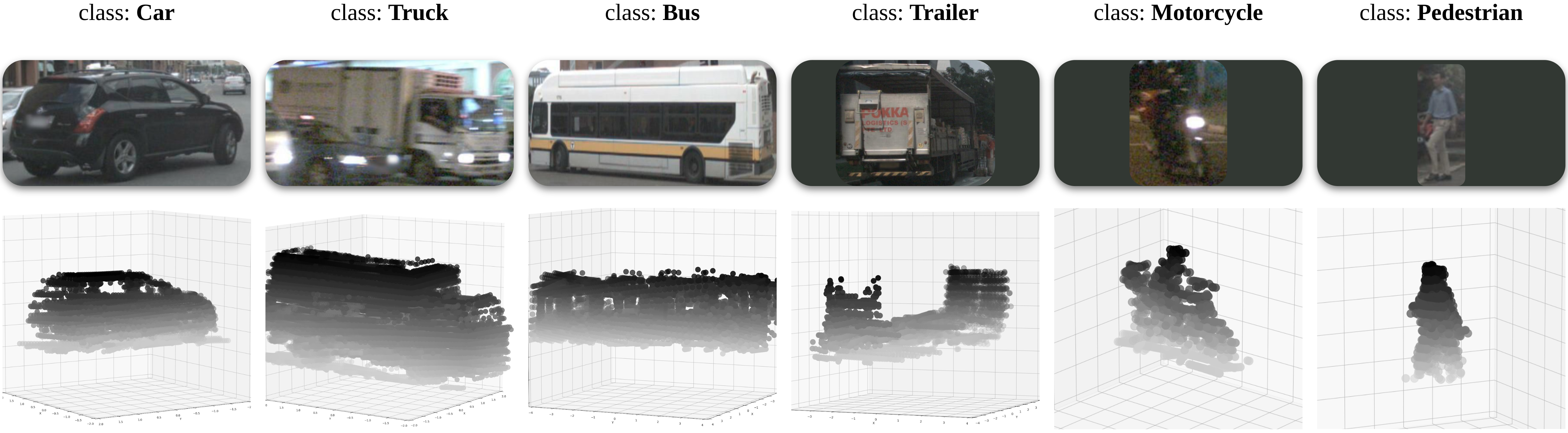}}
    \caption[Sample point clouds of nuScenes dataset object classes]
    {\textbf{Sample point clouds of nuScenes dataset object classes.} In this dataset, the LiDAR point cloud data has been re-annotated from original nuScenes dataset using the BEVFusion C+L algorithm \cite{lidar-reid}. The figure displays camera and LiDAR measurements for various object classes, with the point clouds representing the stacked form of multiple time frames.}
    \label{fig_nuscenes_sample}
\end{figure*}

\subsection{Dataset and Comparison Group}
We evaluate CALM-Net against widely used point-cloud backbones: PointNet, PointNeXt, DGCNN, DeepGCN, and Point Transformer (PTr) \cite{43,44,45,46,47}. These cover direct point processing (PointNet/PointNeXt), graph convolution (DGCNN/DeepGCN), and point attention (PTr), enabling a fair comparison across architectural families.

Training and validation use a nuScenes-based Re-ID dataset containing multiple object classes spanning rigid and deformable categories. To reduce extreme sparsity effects during training, we keep only frames with $>$127 points, yielding seven classes: four rigid (car, truck, bus, trailer), one semi-rigid (motorcycle, often with a rider), one deformable (pedestrian), and one unlabeled class. Sample point clouds for six labeled classes (car, truck, bus, trailer, motorcycle, and pedestrian) are visualized in Fig. \ref{fig_nuscenes_sample}. We use batch sizes of 128 (training) and 256 (evaluation).

\subsection{Re-ID Performance Evaluation}

\begin{table*}[!ht]
\vspace{5px}
\caption[Re-ID performance with random sampling]
{\textbf{Re-ID performance with random sampling (\%)}}
\label{tab_reidcompare}
\begin{center}
\setlength{\tabcolsep}{8.5pt}
\renewcommand{\arraystretch}{1.5}
\begin{tabularx}{\textwidth}{l|ccc|ccccccc}
\hline  
Model & $_{m}$Acc & F1$_{pos.}$ & F1$_{neg.}$ & Acc$_{car}$ & Acc$_{tru.}$ & Acc$_{bus}$ & Acc$_{tra.}$ & Acc$_{mot.}$ & Acc$_{ped.}$ & Acc$_{unl.}$ \\ \hline \hline
PointNet \cite{43} & 92.71 & 92.97 & 92.44 & 94.17 & 93.73 & 85.20 & 89.24 & \underline{87.14} & 80.77 & 92.26 \\
PointNeXt \cite{44} & 92.48 & 92.71 & 92.23 & 93.34 & \underline{94.46} & 87.44 & 88.34 & 81.43 & \textbf{85.38} & 92.18 \\
DGCNN \cite{45} & \underline{93.77} & \underline{93.89} & \underline{93.65} & \underline{95.13} & 94.34 & \underline{88.79} & 88.79 & 82.86 & \underline{85.00} & 93.71 \\
DeepGCN \cite{46} & 93.14 & 93.30 & 92.98 & 94.48 & 94.10 & 87.00 & 89.24 & 85.71 & 82.69 & \underline{96.13} \\
Point Transformer \cite{47} & 93.30 & 93.46 & 93.14 & 94.34 & 94.34 & 86.10 & \underline{90.13} & \textbf{91.43} & \textbf{85.38} & 93.79 \\
\hline
CALM-Net (prop.) & \textbf{94.54} & \textbf{94.72} & \textbf{94.54} & \textbf{95.72} & \textbf{95.57} & \textbf{90.58} & \textbf{91.03} & \underline{87.14} & \textbf{85.38} & \textbf{96.85} \\
\hline
\multicolumn{11}{p{0.95\textwidth}}{\textbf{Bold} text indicates the highest accuracy in each column, while \underline{underlined} text represents the second-highest accuracy. $_{m}$Acc: mean accuracy, \textit{pos.}: positive, \textit{neg.}: negative, Acc$_{*}$: accuracy in class *, \textit{tru.}: truck, \textit{tra.}: trailer, \textit{mot.}: motorcycle, \textit{ped.}: pedestrian, \textit{unl.}: unlabeled, prop.: proposed.}
\end{tabularx}
\end{center}
\vspace{-10px}
\end{table*}

\begin{table*}[!ht]
\caption[Re-ID performance with hybrid sampling]
{\textbf{Re-ID performance with hybrid sampling (\%)}}
\label{tab_reid_fps}
\begin{center}
\setlength{\tabcolsep}{8.5pt}
\renewcommand{\arraystretch}{1.5}
\begin{tabularx}{\textwidth}{l|ccc|ccccccc}
\hline  

Model & $_{m}$Acc & F1$_{pos.}$ & F1$_{neg.}$ & Acc$_{car}$ & Acc$_{tru.}$ & Acc$_{bus}$ & Acc$_{tra.}$ & Acc$_{mot.}$ & Acc$_{ped.}$ & Acc$_{unl.}$ \\ \hline \hline

\multirow{2}{*}{PointNet \cite{43}} & 93.54 & 93.75 & 93.30 & 95.10 & 94.46 & 87.89 & 87.89 & 85.71 & 81.15 & 92.50 \\
& (+0.83) & (+0.78) & (+0.86) & (+0.93) & (+0.73) & (+2.69) & (-1.35) & (-1.43) & (+0.38) & (+0.24) \\ \hline

\multirow{2}{*}{PointNeXt \cite{44}} & 93.04 & 93.26 & 92.82 & 93.68 & 93.59 & \textbf{90.58} & \underline{90.58} & 85.71 & 85.77 & 92.82 \\
& (+0.56) & (+0.55) & (+0.59) & (+0.34) & (-0.87) & (+3.14) & (+2.24) & (+4.28) & (+0.39) & (+0.64) \\ \hline

\multirow{2}{*}{DGCNN \cite{45}} & \underline{94.90} & \underline{95.01} & \underline{94.79} & \underline{96.20} & \textbf{95.94} & 86.55 & \underline{90.58} & \textbf{92.86} & 85.38 & 95.65 \\
& (+1.13) & (+1.12) & (+1.14) & (+1.07) & (+1.60) & (-2.24) & (+1.79) & (+10.00) & (+0.38) & (+1.94) \\ \hline

\multirow{2}{*}{DeepGCN \cite{46}} & 94.06 & 94.17 & 93.95 & 95.21 & 94.71 & \underline{90.13} & 89.69 & 87.14 & 85.38 & \underline{97.66} \\
& (+0.92) & (+0.87) & (+0.97) & (+0.73) & (+0.61) & (+3.13) & (+0.45) & (+1.43) & (+2.69) & (+1.53) \\ \hline

\multirow{2}{*}{Point Transformer \cite{47}} & 93.22 & 93.42 & 93.01 & 94.08 & 95.20 & 81.61 & 89.24 & \underline{91.43} & \textbf{89.23} & 92.58 \\
& (-0.08) & (-0.04) & (-0.13) & (-0.26) & (+0.86) & (-4.49) & (-0.89) & (0.00) & (+3.85) & (-1.21) \\ \hline

\multirow{2}{*}{CALM-Net (prop.)} & \textbf{95.74} & \textbf{95.79} & \textbf{95.69} & \textbf{97.05} & \underline{95.82} & \textbf{90.58} & \textbf{91.03} & 90.00 & \underline{87.69} & \textbf{98.55} \\
& (+1.20) & (+1.07) & (+1.16) & (+1.33) & (+0.25) & (+0.00) & (+0.00) & (+2.86) & (+2.31) & (+1.69) \\ \hline

\multicolumn{11}{p{0.95\textwidth}}{\textbf{Bold} text indicates the highest accuracy in each column, while \underline{underlined} text represents the second-highest accuracy. $_{m}$Acc: mean accuracy, \textit{pos.}: positive, \textit{neg.}: negative, Acc$_{*}$: accuracy in class *, \textit{tru.}: truck, \textit{tra.}: trailer, \textit{mot.}: motorcycle, \textit{ped.}: pedestrian, \textit{unl.}: unlabeled, prop.: proposed. Values in parentheses (·) are expressed in \% points.}

\end{tabularx}
\end{center}
\end{table*}

\begin{figure*}[!ht] 
    \centering
    \includegraphics[width=1.0\linewidth]{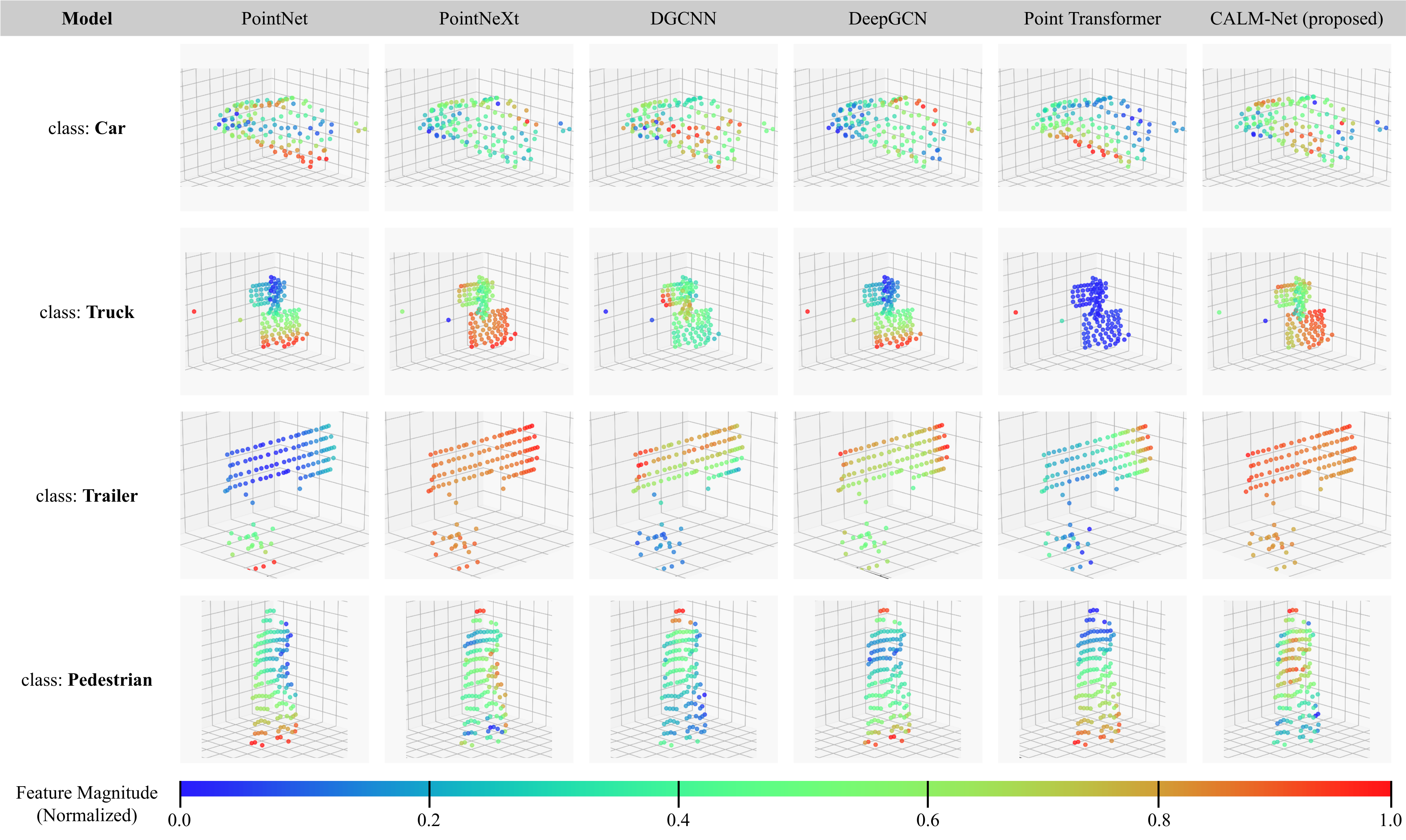}
    \caption[Feature activation heatmaps across different backbone models]
    {\textbf{Feature activation heatmaps across different backbone models.} Feature magnitudes are encoded as colors using a blue–green–red scheme, where blue indicates low activation and red indicates high activation. Each row shows results from a different model, illustrating their learned focus during feature extraction.}
    \label{fig_qual_table}
    \vspace{-2px}
\end{figure*}


\begin{table*}[!ht]
\caption[Complexity analysis of point cloud Re-ID models]
{\textbf{Complexity analysis of point cloud Re-ID models at varying point resolutions.}}
\label{tab:complexity}
\begin{center}
\setlength{\tabcolsep}{8.4pt}
\renewcommand{\arraystretch}{1.4}
\begin{tabularx}{\textwidth}{l|c|c|c|c|c}
\hline
Model & \# Points & Params (M) & Latency (ms) [bs=1/2/4/8] & Throughput (samples/s) [bs=1/2/4/8] & Peak Memory (MB) \\ \hline \hline
PointNet \cite{43} & 128 & 109.34 & 2.49 / 2.58 / 2.76 / 3.28 & 402.28 / 774.56 / 1449.98 / 2440.70 & 430.39 \\
PointNeXt \cite{44} & 128 & 146.73 & 7.42 / 7.22 / 9.20 / 11.25 & 134.80 / 276.83 / 434.71 / 711.35 & 624.67 \\
DGCNN \cite{45} & 128 & 107.15 & 2.42 / 2.64 / 3.39 / 5.14 & 413.01 / 758.25 / 1180.08 / 1554.94 & 449.95 \\
DeepGCN \cite{46} & 128 & 105.32 & 4.00 / 4.21 / 4.29 / 6.42 & 249.75 / 474.65 / 931.37 / 1246.59 & 461.41 \\
Point Transformer \cite{47} & 128 & 219.77 & 21.81 / 21.13 / 21.20 / 21.31 & 45.85 / 94.67 / 188.64 / 375.32 & 858.38 \\
CALM-Net (prop.) & 128 & 220.94 & 22.11 / 22.41 / 26.49 / 24.33 & 45.24 / 89.24 / 150.97 / 328.78 & 901.01 \\ \hline
PointNet \cite{43} & 256 & 109.34 & 2.50 / 2.74 / 3.20 / 4.47 & 400.06 / 728.62 / 1248.40 / 1789.82 & 443.42 \\
PointNeXt \cite{44} & 256 & 146.73 & 7.90 / 8.63 / 11.47 / 17.05 & 126.63 / 231.76 / 348.68 / 469.11 & 678.44 \\
DGCNN \cite{45} & 256 & 107.15 & 2.83 / 3.43 / 5.00 / 8.49 & 353.25 / 582.43 / 800.22 / 941.86 & 491.51 \\
DeepGCN \cite{46} & 256 & 105.32 & 4.07 / 4.57 / 6.82 / 11.00 & 245.88 / 437.81 / 586.62 / 727.36 & 521.16 \\
Point Transformer \cite{47} & 256 & 219.77 & 23.27 / 22.24 / 21.74 / 23.00 & 42.97 / 89.94 / 183.99 / 347.83 & 864.01 \\
CALM-Net (prop.) & 256 & 220.94 & 23.59 / 24.55 / 25.79 / 27.74 & 42.39 / 81.45 / 155.10 / 288.38 & 945.78 \\ \hline
\multicolumn{6}{p{0.95\textwidth}}{bs: batch size, prop.: proposed.}
\end{tabularx}
\end{center}
\vspace{-10px}
\end{table*}


According to Table \ref{tab_reidcompare}, the proposed CALM-Net achieves the highest mean match accuracy, F1 positive score, and F1 negative score among the backbone models. While CALM-Net demonstrates the best performance on most object classes, the motorcycle and pedestrian classes show no improvement, or even slight degradation in match accuracy. This result suggests that CALM-Net is highly effective in re-identifying rigid object classes but less effective for deformable classes such as motorcycles and pedestrians.

Moreover, the adoption of the hybrid subsampling strategy further improved the performance of the model. The comparison between Tables~\ref{tab_reidcompare} and~\ref{tab_reid_fps} shows that adopting the hybrid subsampling strategy (random subsampling during training and FPS during inference) generally improves Re-ID performance across different backbones. Most models exhibit consistent gains in mean accuracy and F1 scores, with the magnitude of improvement varying by architecture and object class. In particular, CALM-Net benefits from FPS at inference, achieving the highest overall accuracy among all tested models. It is also worth noting that certain backbones show slight decreases in specific classes, reflecting the trade-off between random variability and structural fidelity. Nevertheless, the overall trend demonstrates that the hybrid subsampling strategy leads to more robust and reliable Re-ID performance. The second rows in Table~\ref{tab_reid_fps} report the exact change in performance for each model and metric.

To gain deeper insight into how different backbone models attend to object structures, we visualize the learned feature magnitudes overlaid on the input point clouds in Fig.~\ref{fig_qual_table}. Specifically, we examine samples from four object classes—car, truck, trailer, and pedestrian—chosen to represent both rigid and deformable categories. We omit the visualization for buses and motorcycles, as their patterns are qualitatively similar to cars and pedestrians, respectively. Feature magnitudes are mapped to colors using a blue–green–red colormap: blue indicates low feature activation (i.e., less model attention or representational strength), while red indicates high activation, reflecting stronger feature encoding.

Different backbone models yield varying magnitudes of feature activations across the point cloud. In some cases, these models exhibit relatively low activations for specific object classes, meaning that important geometric details may be overlooked. By contrast, CALM-Net tends to produce a more uniformly distributed activation pattern across those same object classes. This indicates that the model is less prone to concentrating excessively on localized regions and instead captures a broader representation of the object surface.

That said, this observation does not imply that CALM-Net universally outperforms all other backbones over all object classes. Other models can also produce strong and well-distributed activations in favorable samples, and if such cases dominate, their Re-ID accuracy may surpass CALM-Net. The key point is that CALM-Net demonstrates a more consistent ability to maintain balanced feature magnitudes, even in scenarios where baseline backbones show weaker activations.

\subsection{Complexity Analysis}

Beyond accuracy, it is essential to assess whether Re-ID models are deployable under real-time constraints, especially in LiDAR point cloud-based perception pipelines. Table~\ref{tab:complexity} provides a detailed comparison of representative point cloud models in terms of parameter count, inference latency, throughput, and GPU memory usage, evaluated across two typical point cloud resolutions (128 and 256 points). These resolutions reflect practical subsampling strategies used in resource-constrained autonomous systems.

The results reveal clear computational trade-offs across backbone architectures. Lightweight networks such as PointNet and DGCNN offer extremely fast inference, achieving latencies as low as 2--3~ms and throughput exceeding 350~samples/s even at 256 points for batch size 1. Models like PointNeXt and DeepGCN provide a middle ground, introducing slightly more overhead (4--8~ms latency). Transformer-based models such as Point Transformer incur significantly higher computational cost due to global attention mechanisms. At 256 points, their latency reaches over 23~ms per frame, with throughput dropping below 43~samples/s in batch size 1. These trends demonstrate a general trade-off: architectures with stronger global reasoning typically sacrifice real-time feasibility unless further optimized.

Despite its multi-branch architecture and curvature-aware embedding module, the proposed CALM-Net maintains competitive efficiency. At 256 input points, it operates at 23.6~ms per frame with batch size 1---well within real-time constraints. This latency translates to approximately 42~samples/s throughput, comfortably exceeding the frame rate of commercial LiDAR sensors, which typically operate at 10--20~Hz (i.e., 50--100~ms intervals). Moreover, CALM-Net’s GPU memory footprint remains under 1~GB, making it suitable for deployment on embedded platforms with limited hardware resources. These results confirm that CALM-Net not only delivers state-of-the-art Re-ID accuracy but also satisfies the latency and memory requirements for real-time deployment in LiDAR-based autonomous systems.

\subsection{Ablation Studies}
The ablation studies were conducted under the same experimental settings as those in Section \ref{sec_experiment}, with the only change being an decrease in the training epochs to 500.

\begin{table}[h]
\caption[MI and Re-ID Performance]
{\textbf{MI and Re-ID Performance}}
\label{tab_miscores}
\centering
\setlength{\tabcolsep}{8.5pt}  
\renewcommand{\arraystretch}{1.5}  
\begin{tabularx}{\columnwidth}{l|>{\centering\arraybackslash}X|
                                 >{\centering\arraybackslash}X 
                                 >{\centering\arraybackslash}X 
                                 >{\centering\arraybackslash}X}  
\hline  
\multicolumn{5}{c}{Baseline Model: Point Transformer} \\
\hline  
Aggregated Model & MI & $_{m}$Acc & F1$_{pos.}$ & F1$_{neg.}$ \\
\hline \hline  
PointNet & 0.19 & 93.57\% & 93.76\% & 93.38\% \\
PointNeXt & 0.19 & 94.02\% & \underline{94.16\%} & 93.88\% \\
DGCNN & \textbf{0.10} & \textbf{94.20\%} & \textbf{94.29\%} & \textbf{94.11\%} \\
DeepGCN & \underline{0.18} & \underline{94.04\%} & \underline{94.16\%} & \underline{93.92\%} \\
\hline
\multicolumn{5}{p{0.95\columnwidth}}{\textbf{Bold} text indicates the best value in each column's evaluation criteria, and \underline{underlined} text indicates the second-best. $_{m}$Acc: mean accuracy, \textit{pos.}: positive, \textit{neg.}: negative.}
\end{tabularx}
\vspace{-10px}
\end{table}

\begin{table}[h]
\caption[Ablation Study on CALM-Net]
{\textbf{Ablation Study on CALM-Net (\%)}}
\centering
\label{tab_ablation_check}
\renewcommand{\arraystretch}{1.5}  
\begin{tabularx}{1.0\columnwidth}{>{\centering\arraybackslash}m{1.8cm} 
                                 >{\centering\arraybackslash}m{0.7cm} 
                                 >{\centering\arraybackslash}m{0.7cm} | 
                                 >{\centering\arraybackslash}m{1.0cm} 
                                 >{\centering\arraybackslash}m{1.0cm} 
                                 >{\centering\arraybackslash}m{1.0cm}}
\hline 
\multicolumn{6}{c}{Baseline Mechanism: Point Transformer (PA)} \\
\hline
DGCNN (EC) & Curv. & FPS &$_{m}$Acc & F1$_{pos.}$ & F1$_{neg.}$ \\
\hline \hline
& & & 91.80 & 92.08 & 91.49 \\
& & V & 92.64 & 92.89 & 92.36 \\
V & & & 94.20 & 94.29 & 94.11 \\
V & & V & \underline{94.75} & \underline{94.84} & 94.65 \\
& V & & 93.73 & 80.77 & \underline{94.84} \\
& V & V & 94.73 & 94.81 & 94.64 \\
V & V & & 94.24 & 94.35 & 94.12 \\
V & V & V & \textbf{95.06} & \textbf{98.71} & \textbf{94.98} \\
\hline
\multicolumn{6}{p{0.95\columnwidth}}{\textbf{Bold} text indicates the highest accuracy in each column, and \underline{underlined} text indicates the second-highest accuracy. Curv: curvature embedding, $_{m}$Acc: mean accuracy, \textit{pos.}: positive, \textit{neg.}: negative.}
\end{tabularx}
\end{table}

To determine the most effective point processing mechanisms for feature aggregation, we adopted an approach based on probability and information theory: mutual information (MI). The MI between two random variables \(X\) and \(Y\) is defined in Eqn. \ref{eq:mi_score}, where \(p(x,y)\) represents their joint probability distribution, and \(p(x)\) and \(p(y)\) denote the marginal probability distributions of \(X\) and \(Y\), respectively.

\begin{equation}\label{eq:mi_score}
\text{MI}(X, Y) = \sum_{x \in X} \sum_{y \in Y} p(x, y) \log\left(\frac{p(x, y)}{p(x)p(y)}\right)
\end{equation}
\vspace{5px}

The primary objective of using MI is to assess the dependency between two variables. A lower MI indicates weaker dependency, whereas a higher MI suggests stronger dependency. In the context of latent feature vectors, a lower MI implies that the feature vectors contain more distinct information from each other, whereas a higher MI suggests a greater degree of shared information. 

As shown in Table \ref{tab_miscores}, we computed the MI between the baseline model, Point Transformer, and other backbone models. The model with lowest MI with Point Transformer is found as DGCNN, which also shows the best Re-ID accuracy as well. Conducting on this finding, the CALM-Net model is designed to combine the point attention and edge convolution mechanisms.

Given this outperforming combination of point attention and edge convolution mechanisms over other combinations, the ablation studies were conducted using point attention as the baseline mechanism to show the effect of each proposed method: feature aggregation with edge convolution, feature aggregation with curvature embedding, and the adoption of the hybrid subsampling strategy. Table \ref{tab_ablation_check} presents the mean match accuracy, F1 positive score, and F1 negative score for each combination.

\section{Conclusion}
In this study, we propose a new point processing model, CALM-ReID, for LiDAR point cloud-based vehicle Re-ID. Our findings can be summarized as follows: (1) we aggregate multiple point processing mechanisms to enhance feature encoding; (2) we introduce a new data representation for point clouds using eigen decomposition to obtain the curvature embedding; and (3) we employ a hybrid subsampling strategy to improve Re-ID performance in practice. By adopting these approaches, we achieved a mean Re-ID accuracy of 95.74\%, which is approximately 1.97\% point higher than the previously best-performing backbone model.

However, achieving optimal accuracy in deformable object classes remains a persistent challenge. Future research could focus on further refining the network model or exploring additional feature extraction mechanisms that can effectively handle both rigid and deformable object classes within a unified network structure. By enhancing the adaptability and robustness of Re-ID models, we can work toward a solution that excels across diverse scenarios, ultimately ensuring more reliable and accurate vehicle Re-ID in varied environments.

\vspace{-10px}

\begin{IEEEbiography}
[{\includegraphics[width=1in,height=1.25in,clip,keepaspectratio]{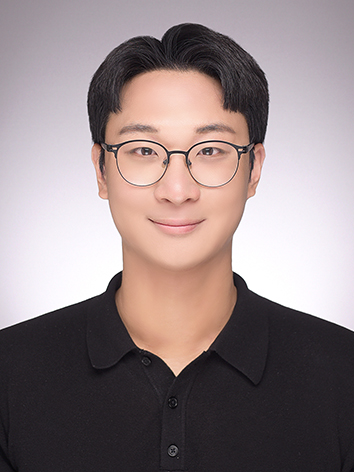}}]
{Dongwook Lee} received the B.Sc. degree from Jacobs University, Bremen, Germany, in Intelligent Mobile Systems, and the M.Sc. degree from Korea Advanced Institute of Science and Technology (KAIST), Daejeon, Republic of Korea in Robotics Program. He is currently pursuing the Ph.D. degree in Robotics Program with KAIST. His research interests include computer vision and autonomous vehicle navigation.
\end{IEEEbiography}
\vspace{-10px}

\begin{IEEEbiography}
[{\includegraphics[width=1in,height=1.25in,clip,keepaspectratio]{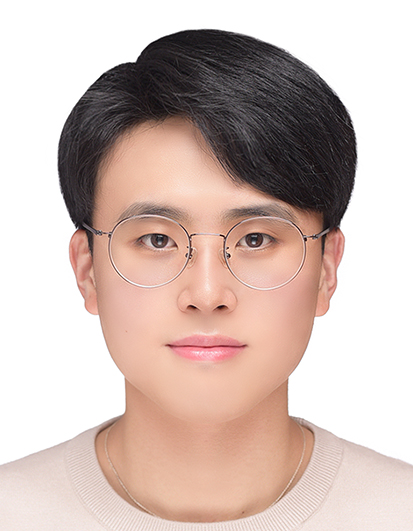}}]
{Sol Han} received the B.S. degree from Korea Advanced Institute of Science and Technology (KAIST), Daejeon, Republic of Korea with a double major in Department of Mechanical Engineering and School of Computing, in 2021. He is currently enrolled in the integrated Master's and Doctoral program in the Department of Mechanical Engineering with KAIST. His research interests include computer vision and sensor fusion.
\end{IEEEbiography}
\vspace{-10px}

\begin{IEEEbiography}
[{\includegraphics[width=1in,height=1.25in,clip,keepaspectratio]{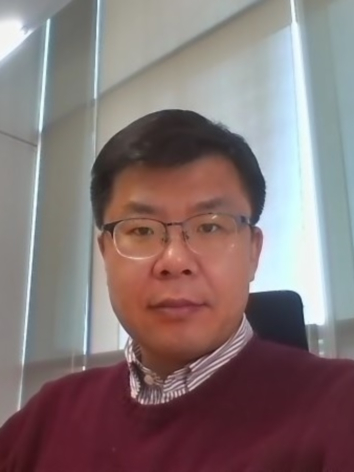}}]
{Jinwhan Kim} received B.S. and M.S. degrees in naval architecture and ocean engineering from Seoul National University, Seoul, Korea and his a Ph.D. in aeronautics and astronautics from Stanford University. Dr. Kim was with Korea Institute of Machinery and Materials, and subsequently with Korea Ocean Research and Development Institute. He is currently with the Department of Mechanical Engineering at KAIST, working in the areas of mobile robotics and control.
\end{IEEEbiography}

\vfill

\end{document}